\documentclass[manuscript,screen]{acmart}

\AtBeginDocument{%
  \providecommand\BibTeX{{%
    \normalfont B\kern-0.5em{\scshape i\kern-0.25em b}\kern-0.8em\TeX}}}

\setcopyright{acmcopyright}
\copyrightyear{2018}
\acmYear{2018}
\acmDOI{XXXXXXX.XXXXXXX}

\acmConference[Conference acronym 'XX]{Make sure to enter the correct
  conference title from your rights confirmation emai}{June 03--05,
  2018}{Woodstock, NY}
\acmBooktitle{Woodstock '18: ACM Symposium on Neural Gaze Detection,
 June 03--05, 2018, Woodstock, NY} 
\acmPrice{15.00}
\acmISBN{978-1-4503-XXXX-X/18/06}

\begin{document}

\title{A Taxonomy of Rater Disagreements: Surveying Challenges \& Opportunities from the Perspective of Annotating Online Toxicity}


\author{Wenbo Zhang}
\affiliation{%
  \institution{Pennsylvania State University}
  \country{USA}}
\email{wjz5120@psu.edu}

\author{Hangzhi Guo}
\affiliation{%
  \institution{Pennsylvania State University}
  \country{USA}}
\email{hangz@psu.edu}

\author{Ian D Kivlichan}
\affiliation{%
  \institution{OpenAI}
  \country{USA}}
\email{idk@openai.com}

\author{Vinodkumar Prabhakaran}
\affiliation{%
  \institution{Google}
  \country{USA}}
\email{vinodkpg@google.com}

\author{Davis Yadav}
\affiliation{%
 \institution{Pennsylvania State University}
 \country{USA}}
 \email{davisyadav@psu.edu}

\author{Amulya Yadav}
\affiliation{%
  \institution{Pennsylvania State University}
  \country{USA}}
\email{amulya@psu.edu}
\renewcommand{\shortauthors}{Zhang, et al.}

\begin{abstract}
  Toxicity is an increasingly common and severe issue in online spaces. Consequently, a rich line of machine learning research over the past decade has focused on computationally detecting and mitigating online toxicity. These efforts crucially rely on human-annotated datasets that identify toxic content of various kinds in social media texts. However, such annotations historically yield low inter-rater agreement, which was often dealt with by taking the majority vote or other such approaches to arrive at a single ground truth label. Recent research has pointed out the importance of accounting for the subjective nature of this task when building and utilizing these datasets, and this has triggered work on analyzing and better understanding rater disagreements, and how they could be effectively incorporated into the machine learning developmental pipeline. While these efforts are filling an important gap, there is a lack of a broader framework about the root causes of rater disagreement, and therefore, we situate this work within that broader landscape. In this survey paper, we analyze a broad set of literature on the reasons behind rater disagreements focusing on online toxicity, and propose a detailed taxonomy for the same. Further, we summarize and discuss the potential solutions targeting each reason for disagreement. We also discuss several open issues, which could promote the future development of online toxicity research. 
\end{abstract}

\ccsdesc[500]{Human-centered computing~Collaborative and social computing theory, concepts and paradigms}
\ccsdesc[300]{Applied computing~Law, social and behavioral sciences}


\keywords{Online toxicity, Inter-rater disagreement, Crowdsourcing, Taxonomy}



\maketitle

\section{Introduction}
\label{sec:intro}
Social network platforms have created a vibrant environment where individuals from various countries, cultures, and backgrounds may exchange ideas and thoughts. These platforms have become a potent force for influencing conversation and public opinion, as they provide a convenient means for news (and other information) to be shared, enabling people to express their ideas and thoughts on public issues in front of a global audience. Unfortunately, the rampant proliferation of social network usage comes with certain negative externalities, e.g., toxicity associated with online discussion (including several subtypes such as cyberbullying, hate speech, profanity-laden and abusive messages, trolling) has become an increasingly common and severe issue in online spaces~\citep{garg2022handling}. The definition of online toxicity can vary depending on researchers in multiple research areas. In this paper, we follow the definition of online toxicity as “comments on online platforms that are rude, disrespectful, or otherwise likely to make someone leave a discussion”~\citep{dixon2018measuring}. It can take many different forms, e.g., insulting hints to actions that move from the web to the offline world, thus posing a threat to people's safety and lives~\citep{babakov2022beyond}. Thus, it is crucial to detect and mitigate such kinds of toxic communication in online spaces.

Over the last decade, there have been lots of efforts from academic and industrial researchers aimed at building computational machine learning models to detect and mitigate online toxicity on social media platforms. However, it remains challenging to build such models as they rely on annotated datasets that contain accurate text-label pairs where the label (or annotation) denotes the presence (or absence) of toxicity in the text. More precisely, annotation is the process of providing metadata (e.g., deeper meaning, context, nuance) through the act of labeling language~\citep{DBLP:conf/hicss/PattonBFGK19}. For the topic of comment toxicity discussed in this paper, the process of the annotation involves assigning different types of toxicity labels to text comments (these comments may take the form of phrases, sentences, and/or paragraphs). Quite often, these annotated datasets are created with the help of annotations provided by professional annotators or raters from crowdsourcing platforms such as Amazon Mechanical Turk, Crowdspring, 99designs, etc.

Most prior research in this area assumes that for a given comment (or data point), there is one golden "ground-truth" (or correct) toxicity label. Hence, it is standard practice in crowdsourcing literature to get annotations from multiple raters for each data point, and it is hoped that taking the majority vote of the annotations provided by multiple raters will lead us to the ground-truth label. However, such annotations historically yield low inter-rater agreement, and while techniques have been developed to reduce inter-rater disagreement (thereby improving the accuracy of majority-vote annotations)~\citep{miltsakaki-etal-2004-annotating}, these approaches rarely shed light on the reasons behind rater disagreements, and how understanding these reasons can enable us to develop more sophisticated and nuanced tools for dealing with online toxicity.


Evaluating online comments for signs of toxicity is an intrinsically subjective task for raters\footnote{We will use raters to represent workers or annotators in the annotation task unless explicitly specified}, as their interpretation of toxicity depends on multiple factors, such as background, expertise, and experiences \cite{denton2021whose}. As such, there may not even exist any ground-truth (or correct) label which determines the presence (or absence) of toxicity \cite{aroyo2015truth}. Thus, in this domain, the proportion of raters who annotate certain comments as toxic may simply capture the likelihood of that comment being toxic, rather than grounding it in a singular ground truth of its toxicity. In this paper, we argue that rater disagreement is core to the toxicity annotation task, and further, it can sometimes be important to explicitly capture certain types of rater disagreement: what in other contexts might be considered “label noise” is essential in our domain.


Unfortunately, distinguishing rater disagreements that we want to capture -- broadly speaking, disagreements that arise out of differing subjective beliefs and opinions -- from other types of unintended disagreements, such as noise or incorrect judgments is not straightforward. 
These unintended disagreements induce a gap between what we model (the probability that the rater, given limited time/attention, finds a comment toxic) from what we intend to model (the probability that an average reader would find that comment toxic). More formally, we can view these undesired and desired disagreements as reducible and irreducible sources of noise, respectively; i.e.\ as we provide more information to raters -- missing context, subject matter expertise, judgment time, etc. -- the former decreases, while the latter remains constant.

In this survey paper, we develop a comprehensive taxonomy of the reasons for rater disagreements focused on the problem of online toxicity. For each type of rater disagreement, we include a loose discussion including example causes, and (for unintended types of rater disagreement) approaches to mitigating them. This taxonomy aims to be comprehensive by capturing both intended (or intrinsic) as well as unintended sources of disagreement. In the longer term, this taxonomy serves as a stepping stone for the development of more general methods for separating and disentangling the intrinsic sources of rater disagreement from the unintended types of rater disagreement and label noise. The remainder of this survey is organized as follows: Section 2 discusses related work on rater disagreements. Section 3 analyzes different types of reasons behind inter-rater disagreement in the context of toxicity annotation, and proposes a comprehensive taxonomy of these reasons. Section 4 summarizes approaches from previous works about how to deal with different types of disagreements. For example, we discuss approaches that can be used to eliminate or even utilize disagreements to obtain better annotation quality. Section 5 discusses the open research directions for the crowdsourcing and AI research community to deal with toxicity and rater disagreement. We conclude the paper by summarizing major findings and discussing remaining issues for future work.

\section{Related Work on Rater Disagreements}
\label{sec:reasons}
Recent work has discussed reasons behind inter-rater disagreements under different scenarios. For example, ~\citet{basile-etal-2021-need} focused on several subjective tasks in natural language processing (NLP) and computer vision, and discussed the source of inter-rater disagreement from the vantage points of the rater, the data, and the context. Based on the ``Triangle of Reference'', \citet{aroyo2015truth} developed another taxonomy of disagreement for NLP-related tasks that highlighted reasons such as uncertainty in the meaning of sentences, under-specification of guidelines for annotation, and rater behavior. Similarly, ~\citet{jiang2022investigating} investigated rater disagreements in the area of natural language inference. To the best of our knowledge,~\citet{uma2021learning} is the survey paper closest to the theme of our paper, but it focuses on several natural language processing and computer vision tasks, instead of online toxicity detection. It classifies the reasons for rater disagreement into five categories: (1) rater errors and interface problems; (2) incomplete or vague annotation schemes; (3) ambiguity (expressions can be interpreted in semantically distinct ways in a given context~\citep{poesio2020ambiguity}); (4) item difficulty (the interpretation of an item is hard to define); (5) rater's subjectivity. Similarly, ~\citet{sandri2023don} also proposed a taxonomy of reasons for rater disagreements into four categories: (1) sloppy annotation (issues related to carelessness of raters); (2) ambiguity; (3) missing information (to interpret the text content properly); (4) rater's subjectivity including identity, beliefs, and background. Unfortunately, no prior work in this space focuses on the domain of online toxicity, and hence, we address this research gap by proposing a comprehensive taxonomy of the reasons behind inter-rater disagreement for this domain. Our taxonomy builds upon the knowledge derived from taxonomies proposed in prior work \cite{basile-etal-2021-need,aroyo2015truth,uma2021learning}, and we add to that knowledge base in our work by introducing several new reasons for inter-rater disagreement (each of which is further broken down into several sub-categories), with a strong focus on the online toxicity domain (which has not been explored in prior work).  

\noindent \textbf{Survey Design.} We now describe our survey design process, i.e., we provide a brief discussion on (i) the methods used for the identification of articles that could potentially be included in the survey, (ii) screening the articles for review, (iii) deciding on the articles eligibility for our survey paper, and eventually (iv) finalizing the list of articles that we included in our survey paper.

To begin with, we used the five top-level reasons for rater disagreement identified by the taxonomy proposed by~\citet{uma2021learning} as a starting point for our research. Based on the five reasons outlined in~\citet{uma2021learning}, we used a series of keywords such as ``rater errors", ``incomplete annotation", ``ambiguity in annotation", ``rater subjectivity", ``toxicity annotation", etc., and used different subsets of these keywords to query Google Scholar. This enabled us to identify an initial list of papers. Next, the title and abstracts of these identified papers were manually searched for keywords such as ``toxicity annotation" and ``rater disagreements", etc., to screen out irrelevant papers returned by our Google Scholar queries. Next, for a paper to be considered eligible for inclusion in our survey paper, it had to be written in English language, and a preference was given to papers published in recent proceedings of relevant human computation, NLP-related conferences, journals, and workshops. We also check the citing and cited papers of the eligible papers to further expand our list of relevant papers. This gave us the final list of papers related to different reasons for inter-rater disagreement that we have chosen to include in Sections 3 and 4 of our survey paper.


\section{Reasons for Inter-Rater Disagreement}

\begin{figure*}
  \centering
  \includegraphics[width=\linewidth]{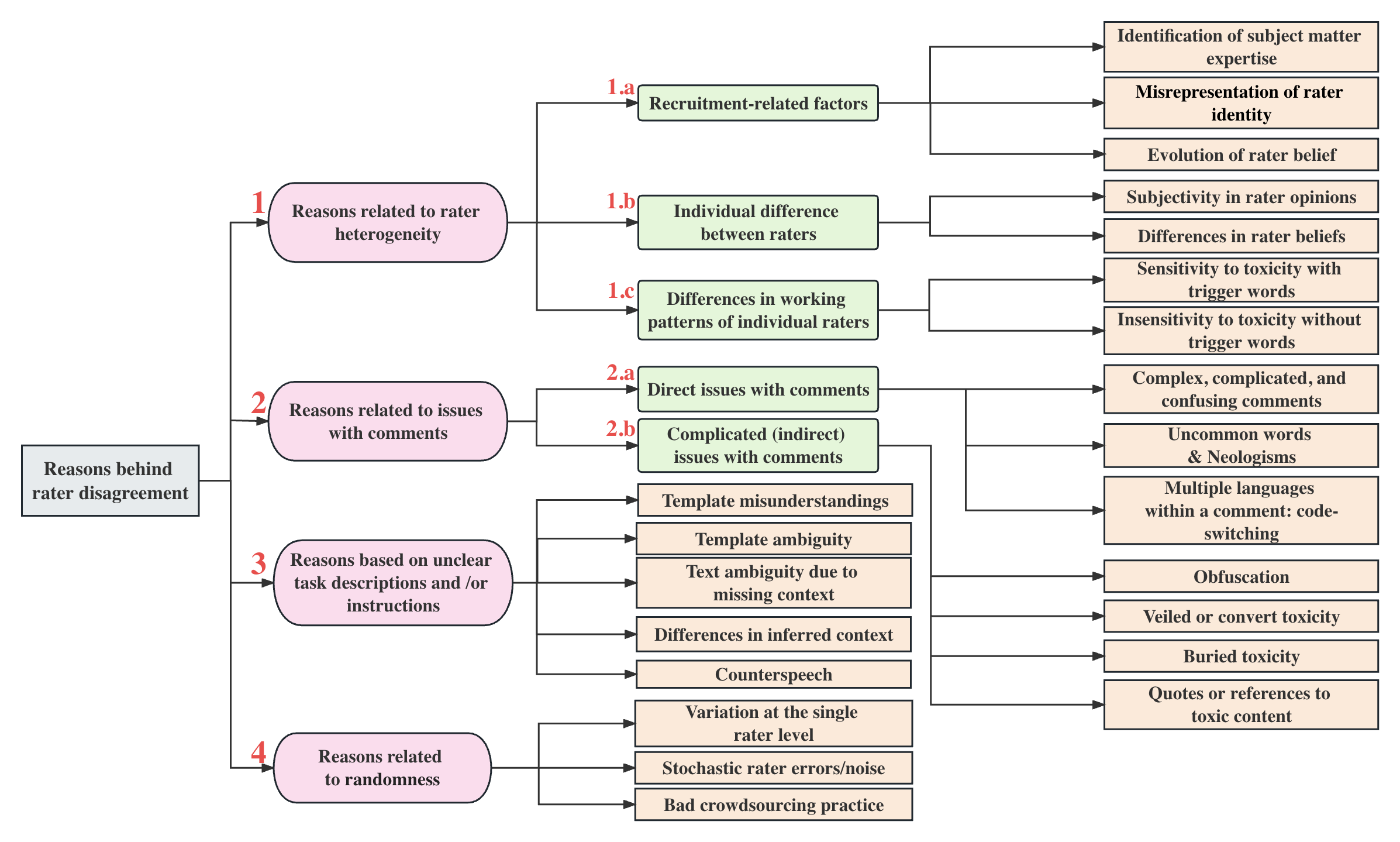}
  \caption{Our proposed taxonomy of reasons behind rater disagreements focusing on online toxicity annotation, where boxes in pink represent top-level reasons (1-4), boxes in green represent intermediate-level reasons (1.a-1.c \& 2.a-2.b), and boxes in orange represent bottom-level reasons.}
  \label{fig:comprehensive_taxonomy}
\end{figure*}

\begin{figure*}
  \centering
  \includegraphics[width=\linewidth]{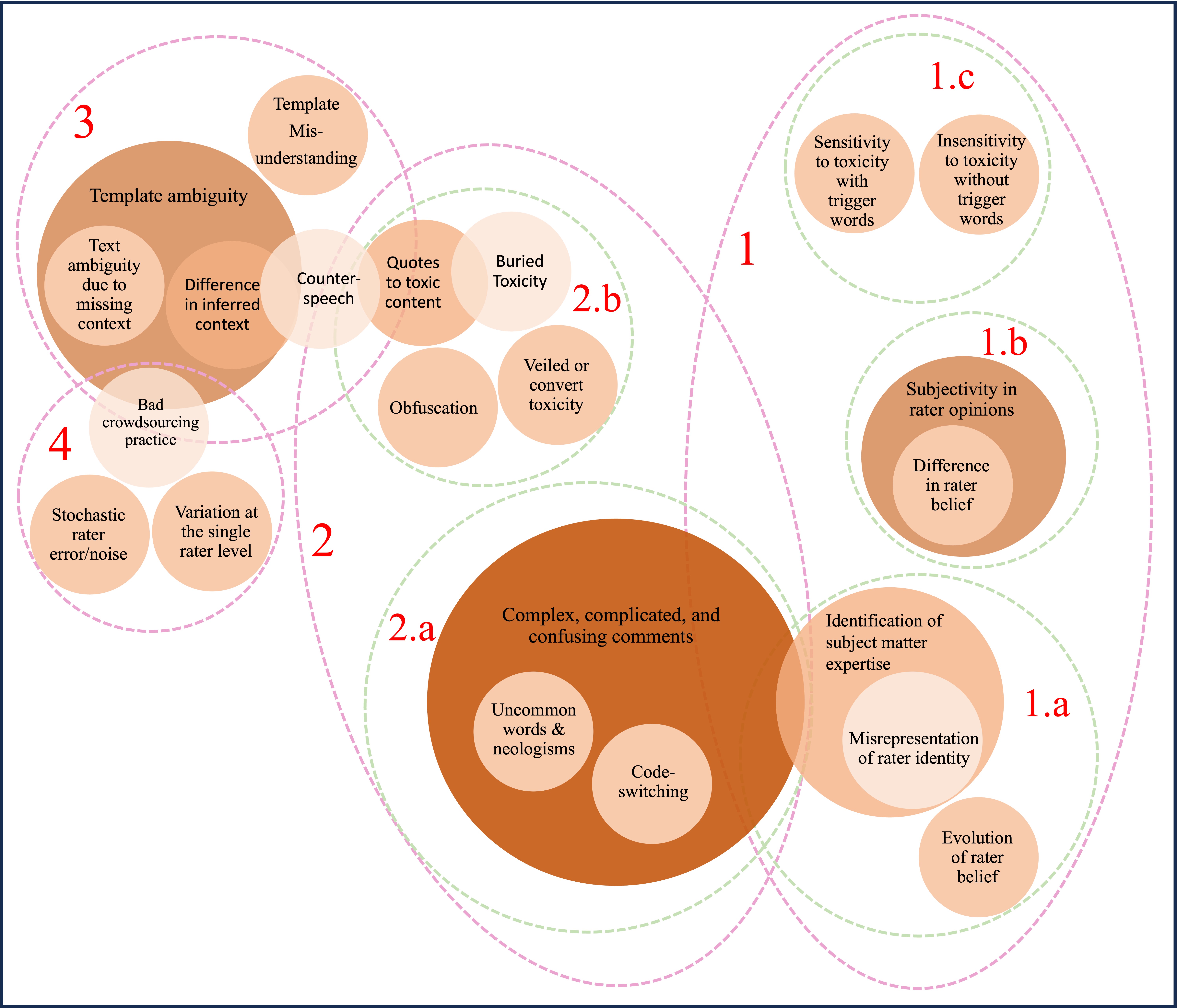}

  \caption{Venn diagram representation of our proposed taxonomy of reasons behind rater disagreements focusing on online toxicity annotation that illustrates subset and overlapping relationships between different reasons for rater disagreement. Similar to the coloring scheme used in Figure 1, the pink dotted ovals represent top-level reasons (1-4), the green dotted ovals represent intermediate-level reasons (1.a-1.c \& 2.a-2.b), and ovals with shades of orange represent the various bottom level reasons outlined in Figure 1.}
  \label{fig:venn_diagram}
\end{figure*}

The scope of this paper is focused on reasons for rater disagreements during the annotation procedure of online toxicity. As mentioned before, the annotation of toxic comments is an intrinsically subjective task. There could be various reasons for rater disagreements in the toxicity annotation process, depending on the particular context. In this section, we try to discuss all possible reasons behind rater disagreement. We categorize all these reasons into a comprehensive taxonomy of reasons behind rater disagreements, which is illustrated in Figures~\ref{fig:comprehensive_taxonomy} \&~\ref{fig:venn_diagram}. Broadly, we generalize all possible reasons into four top-level categories, including (i) reasons related to rater heterogeneity, (ii) reasons related to issues with text comments (which need to be annotated), (iii) reasons related to unclear task descriptions, and (iv) reasons arising from ``true" randomness. Further, we subdivide these four high-level categories by proposing subtypes that take into account distinct real-world scenarios.

\subsection{Reasons related to Rater Heterogeneity}
\label{sec:heterogeneity}
Heterogeneity among raters can lead to variations in their ability to annotate or classify toxic comments. For example, differences in language proficiency and skillsets can affect how raters understand and interpret the comments being annotated. Raters with lower proficiency in the language used in the comment may struggle to accurately identify and classify toxic behavior. Another possible reason is variations in the subjective interpretation of toxic comments~\citep{10.1145/3308560.3317083}. For example, one rater may consider a comment to be toxic while another may perceive it to be merely critical or sarcastic. From another perspective, variations in cultural backgrounds and experiences can also influence how raters interpret and assess toxicity~\citep{kumar2021designing}. For example, what is considered toxic in one culture may not be perceived as such in another, leading to inconsistencies in annotations. Finally, variations in the level of training and experience can also affect the quality and consistency of raters' annotations~\citep{waseem-2016-racist}. Experienced raters may have developed a more nuanced understanding of toxicity and thus, they may be better equipped to distinguish between different types of toxic behavior. In this section, we explore reasons related to heterogeneity from three perspectives: recruitment-related factors, individual differences between raters, and differences in the working patterns of individual raters.\\

\noindent\textit{3.1.1 Recruitment-related Factors}\\
The first subset of reasons behind inter-rater disagreement during the annotation of online toxicity pertains to issues that arise during the recruitment of raters for the task of annotation. For example, crowdsourcing platforms and job requesters (who publish their annotation tasks on these platforms) may lack the means to identify and recruit raters with sufficient background and expertise to finish a particular annotation task. An additional layer of nuance is added by the fact that different annotation tasks may require different kinds of expertise (possessed by different raters), and it is very difficult for job requesters to find the right raters for their annotation task. Below, we discuss three specific types of recruitment-related factors that lead to inter-rater disagreement.\\

\textbf{Challenges in Identifying Subject Matter Experts.} As briefly discussed above, the first scenario that generates disagreements is that comments assigned to raters require specific kinds of expertise (e.g.,\ the identity-derived lived experiences found in specialized rater pools, or cultural contexts). In fact, ~\citet{8554954} indicated that the nationality of a rater significantly affected their ratings of hatefulness on social media comments. In their study, the authors sampled crowd workers from 50 countries and let them score the same social media comments for toxicity. Experimental results indicated that the perceived hatefulness of the comments varied by the users’ subjective experiences. Similarly, ~\citet{al-kuwatly-etal-2020-identifying} demonstrated how particular demographic features of raters, such as first language, age, and education, might bias their annotations for datasets on hate speech classification. Further, ~\citet{kumar2021designing} also concluded that a rater’s attitude toward filtering toxic content was moderated by a multitude of factors, such as demographic background, their personal experiences with harassment, and even their attitudes towards technology and the state of toxic content online. Based on experiments from widely-used corpora of annotated toxic language, the rater’s social identity (e.g., race) was also found as a moderating factor that leads to rater disagreement during the annotation procedure~\citep{sap-etal-2019-risk}. A similar conclusion is also obtained by~\citet{goyal2022toxicity}, where rater identity is a statistically significant factor in how raters will annotate toxicity for identity-related annotations.~\citet{sang2022origin} explored why such disagreements occur in subjective data labeling tasks such as hate speech annotation, and showed that personality and age had a substantial influence on the dimensional labeling of hate speech.

Unfortunately, while existing crowdsourcing platforms enable job requesters to filter raters on the basis of several qualifications (e.g., age, stock ownership, annual income, etc., are premium qualifications for recruitment of raters that are available to job requesters on Amazon Mechanical Turk for additional fees), there are three challenges: (i) this list of premium qualifications for filtering potential raters afforded by crowdsourcing platforms is not (and perhaps, can never be) exhaustive; (ii) more importantly, job requesters may themselves not have enough knowledge about what rater characteristics would be best equipped to tackle their specific annotation task (e.g., it is unclear which raters would be ideal for annotating data related to ethical dilemmas in the Moral Machine experiment \cite{awad2018moral}); and finally, (iii) these ideal rater characteristics may vary with the characteristics of the annotation task, which further complicates the task of identifying subject matter experts or ideal raters.\\

\textbf{Challenges due to Misrepresented Raters.} The challenge of identifying subject matter experts (or ideal raters) is further complicated by the fact that many raters on crowdsourcing platforms commonly misrepresent themselves, or a part of their lived experiences (so as to make themselves eligible for a wider set of tasks on crowdsourcing platforms). For example, raters from outside the US may sometimes represent themselves as English (US) locale to get higher rates, but end up misunderstanding locale-specific terms while annotating text-based comments. As shown in Figure \ref{fig:venn_diagram}, these scenarios correspond to a sub-category of the challenges in \textit{identifying subject matter experts}.

This intentional misrepresentation of identities further leads to inaccurate (or poor) recruitment, which leads to disagreements in toxicity annotation tasks. ~\citet{DBLP:conf/hicss/PattonBFGK19} underscored the need for domain expertise when reviewing Twitter data from vulnerable populations.~\citet{waseem-2016-racist} pointed out the importance of recruiting expert raters for the hate speech annotation task and found that amateur raters were more likely than expert raters to (wrongly) label items as hate speech. The corresponding experiment shows that models trained on expert annotations outperform similar models trained on amateur annotations. 

Unfortunately, existing crowdsourcing platforms currently do not have the means to prevent raters from misrepresenting their identities. For example, Amazon Mechanical Turk (AMT) relies on monitoring browser and client-specific metadata as a means to check for signs of any suspicious activities from raters; unfortunately, it is very common for raters belonging to countries in the Global South to use Virtual Private Networks (VPNs) to create accounts on AMT with misrepresented identities \cite{downs2010your}.\\

\textbf{Challenges due to Dynamically Evolving Rater Beliefs.} A third layer of recruitment-related reasons for rater disagreement arises from the fact that rater beliefs change over time. For instance, even if the same rater is asked to rate the same comment multiple times over a sufficiently long time period, we may observe disagreement among the multiple annotations provided by the same rater, and this disagreement in truth would reflect an evolving understanding and belief system of the rater. In fact, ~\citet{10.1145/3415163} showed the variation tendency of the intensity of hate comments on online platforms over time, and the results illustrated the temporal shift in datasets corresponding to hateful comments. Machine learning models trained on datasets by such raters will struggle to update to the new/correct labels without re-annotation, though this change will be more natural for human raters.\\

\noindent\textit{3.1.2 Individual  Differences between Raters}

\noindent Beyond recruitment-related factors for rater disagreement, there could also be individual-level factors (e.g., unique individual preferences that may not be directly associated with any larger socio-demographic categories that the rater belongs to) that could lead to disagreement between rater annotations. For example, variations in different raters' personal standards (and subjective opinions) for what constitutes toxic (or non-toxic) behavior could impact the differences in their annotations, i.e., one rater may consider a particular behavior to be toxic based on their personal lived experiences, while another may not, leading to variations in their ratings. Below, we discuss two specific types of individual-level factors that lead to inter-rater disagreement.\\


\textbf{Subjectivity in Rater Opinions.} In many situations, it is possible for raters to have the same level of background knowledge and subject matter expertise, and yet, they may disagree in their ratings. These inter-rater disagreements are driven by differences in personal opinions that are shaped by their unique lived experiences, rather than differences in knowledge (i.e.,\ without any implication of error on the part of the raters). While it is desirable to explicitly capture such disagreements arising out of differing opinions (instead of treating them as reducible noise), and machine learning models are intended to account for this subjectivity directly, it is often difficult to disentangle this subjectivity from other types of rater disagreement.~\citet{aroyo2019crowdsourcing} indicated that this type of rater subjectivity is inherent to the topic of comments (i.e.\ topics requiring judgments based on personal preferences or experiences, where two individuals may simply have different opinions).\\

\textbf{Differences in Rater Beliefs.} Raters may be less likely to mark a comment as toxic depending on whether they agree with the ideas expressed in the comment or not. For example, ``\textless rater's favorite sports team\textgreater suck'' may be considered toxic by the rater while ``\textless opposing team\textgreater are stupid'' may not. In fact,~\citet{aroyo2019crowdsourcing} stated that two raters could have different levels of sensitivity to profanity in toxicity annotation tasks, depending on their backgrounds and personal beliefs. As shown in Figure \ref{fig:venn_diagram}, this is a subtype of \textit{rater opinion subjectivity}, but importantly reflects a potentially undesirable type of subjectivity.\\

\noindent\textit{3.1.3 Differences in Working Patterns of Individual Raters}

\noindent In addition to the reasons related to recruitment-related and individual-level factors that lead to rater disagreements, a third type of inter-rater disagreement arises due to differences in the working patterns (or styles) employed by individual raters to accomplish the task of providing annotations for online toxicity. For example, some raters may try to simplify the procedure of annotation (in an attempt to reduce working time and earn money more quickly) by identifying a set of trigger words, and the presence (or absence) of these trigger words can be used as an easy-to-use ``imperfect" heuristic to quickly identify if comments are toxic (or not). We discuss two specific types of working patterns (or heuristics) used by individual raters.\\

\textbf{Sensitivity to Toxicity with Trigger Words.} As a heuristic, raters may rely on the presence of specific trigger words (e.g.,\ curse words) as a way of quickly scoring a comment as toxic, even when they might not consider that comment toxic if they read it more thoroughly. Note that this way of scoring comments is not necessarily a time-saving heuristic employed by raters, as some raters may truly consider certain triggering words toxic in general. Unfortunately, machine learning-based toxicity detection models often face the same issues, i.e., such models rely on decision rules that fixate heavily on the presence of certain trigger words, perhaps in part due to being trained on datasets that are affected by rater sensitivity to trigger word toxicity.  

This sensitivity affects the different subtypes of toxicity in different ways, e.g.\ extreme types of toxicity (e.g., obscenity or profanity-laden comments) often includes more triggering words, and hence, this sensitivity to trigger word toxicity among raters often leads to lesser subjectivity in their annotations. On the other hand, it is less clear how this generalizes to other subtypes of toxicity like threats, trolling, etc. 

Unfortunately, this sensitivity is further complicated by the fact that trigger words of toxicity might differ by communities and by topics due to different norms and usage of language \cite{10.1145/3366423.3380074}. For example, ~\citet{chong2022understanding} analyzed different scenarios (between Singapore and New York) and concluded that toxicity triggers can be tremendously different between Western and Eastern environments. As a result, it would be worthwhile to understand and compare these differences between toxicity triggers to gain deeper insights on online toxicity~\citep{chong2022understanding}.\\

\textbf{Insensitivity to Toxicity without Trigger Words.} Conversely, raters may not consider comments toxic because they do not contain any profanity-laden words or phrases. For example, comments that are abusive or extremely negative in tone, but do not contain explicit slurs or curse words may not be deemed as toxic by raters who are using the presence of trigger words as an inaccurate signal for identifying toxic content. This subtype of insensitivity roughly pairs with sensitivity to trigger words, i.e.,\ raters that use both these heuristics accomplish toxicity annotation tasks by a keyword-matching procedure, instead of reading the comments in detail.  

Once again, machine learning-based toxicity detection models face similar issues. In fact,~\citet{van-aken-etal-2018-challenges} analyzed some of the error cases generated by machine learning models. In particular, they found that some of the false negative cases that machine learning models failed to detect as toxic comments contained toxicity without swear words. For example, “She looks like a horse” was wrongly predicted to be non-toxic by ML models. On the contrary, the occurrence of swear words might not always imply toxicity, and ML models wrongly flagged such instances as false positives. For example, “Oh, I feel like such an asshole now. Sorry, bud.” was a false positive returned by their model. 

\subsection{Reasons related to Issues with Comments}
\label{sec:comment}
While Section \ref{sec:heterogeneity} focused on reasons for rater disagreement that were related to ``issues" induced by raters themselves, we now turn our attention to a completely different set of reasons for rater disagreement, those that relate to issues with the text (or comments) that needs to be annotated. We divide this set of reasons into two broad categories: (i) direct issues with comments; and (ii) complicated (indirect) issues with comments.\\

\noindent\textit{3.2.1 Direct Issues with Comments}

\noindent In this section, we discuss several reasons related to direct issues (or obvious problems) with the comments themselves. For example, it may be the case that raters cannot fully understand the meaning of a comment due to rare, confusing, or unknown words being used inside them. This can result in inconsistencies in toxicity scores and undermine the quality of the resulting annotation. Below, we break down each of these issues separately.\\

\textbf{Complex, complicated, and confusing comments.} As briefly discussed above, comments that are difficult to parse may lead to rater disagreement, as different raters may interpret the confusing comment in different (yet, inaccurate) ways. This is a very common problem with datasets on online toxicity, as the nature of social media comments is often unstructured, and does not follow any specific standards. For example, comments containing abuses and slangs are often difficult to parse, as online commenters commonly express emotion (e.g., waves of anger and depression) through intentional misspellings and repetitive use of adjacent characters~\citep{9290710}. Further, ~\citet{aroyo2019crowdsourcing} also pointed out that subjective text assessment tasks are highly prone to disagreement between judges (or annotators), and this disagreement is typically caused by the ambiguity of the text. These types of comments are potentially difficult for ML models to detect as well as for human raters to annotate.\\


\textbf{Uncommon words \& Neologisms.} Comments containing uncommon (or rare) words that are not known to many raters can also be a cause of rater disagreement.~\citet{10.1145/3078714.3078715} highlighted the importance of task clarity in crowdsourcing, where their results suggest that usage of long words and non-standard languages decreased the task clarity and increased inter-rater disagreement. 

On the other hand, neologisms refer to comments containing completely new words that are not known to all raters. As a result, they may be scored differently by different raters depending on their knowledge and awareness. 

These issues are especially important when they intersect with the previously discussed reason called dynamically evolving rater beliefs (see Section 3.1.1) – for example, if the same rater in the near future would find the comment toxic (as the neologisms or uncommon words become more common), it’s especially important to flag these comments correctly. As shown in Figure \ref{fig:venn_diagram}, uncommon words and neologisms are an important subtype of \textit{complex, complicated, and confusing comments}. Further, this figure shows that uncommon words and neologisms can also be viewed as an instance of the difficulties faced in \textit{identifying subject matter experts} (Section 3.1.1) who would be more knowledgeable about these uncommon words and neologisms.\\


\textbf{Multiple languages within a comment: Code-Switching.} Finally, raters may disagree in their annotation of comments that use a word or phrase from another language that raters are unfamiliar with. For example,~\citet{van-aken-etal-2018-challenges} found that the multilingual nature of online toxicity was one of the three main challenges for toxic comment analysis. We anticipate this to be an increasingly important challenge in the near future, as social media platforms rapidly find new users from the Global South, who often have a very diverse multi-lingual background. As shown in Figure \ref{fig:venn_diagram}, code-switching comments are an important subtype of \textit{complex, complicated, and confusing comments}.
\\

\noindent\textit{3.2.2 Complicated (indirect) Issues with Comments}

\noindent We now discuss reasons for disagreement related to complicated, indirect, or nuanced issues with comments. While direct issues with comments (Section 3.2.1) referred to surface-level (or syntax based) issues, there are a deeper set of semantic-level issues that can arise in situations where raters can clearly check and understand the meaning of each word in the comment directly, yet they disagree in their annotations. This happens because this subtype of toxicity is hidden either in the ``\textit{corners of the text}", or in citations related to the comment piece. We discuss four subtypes of complicated issues with comments. \\

\textbf{Obfuscation.} In many situations, comments are written with obfuscated words, i.e., words rewritten in unusual ways, or in ways that require in-group context to properly moderate. For example, leetspeak may not be interpretable to all raters, and might instead be rated as gibberish  \cite{schiel2012impact}. Hence, annotating comments containing leetspeak is a non-trivial task for both humans and machines. In addition,~\citet{lees-etal-2021-capturing} also discuss that obfuscation could be used to hide toxicity via intentional misspellings, coded words, or implied references. In such situations, it may be the case that different raters interpret obfuscated comments differently, which leads to inter-rater disagreement.\\

\textbf{Veiled or Covert Toxicity.} Quite commonly, the toxicity in comments can be veiled (or hidden) in different ways, e.g., the discussion of a person's characteristics in a disrespectful way, unfair generalizations or stereotyping based on ethnicity, religion, gender, race, etc. For example,~\citet{wiegand-etal-2021-implicitly-abusive} argues that there are different subtypes of implicit abusive language usage and while some of them are obvious in available datasets, others are sparsely found (e.g.,\ dehumanization or euphemisms). Comments containing veiled toxicity are offensive and frustrating, which may lead to inter-rater disagreements~\citep{babakov2022plain}, as raters might have divergent sensitivities and interpretations about what is considered veiled (or covert) toxicity, etc.\\

\textbf{Buried Toxicity.} This kind of toxicity refers to long comments containing a small toxic section which renders the entire comment toxic, but that small toxic section may be missed by raters quickly reading the text. This is different from veiled toxicity in that the writer (or commenter) does not attempt to hide it; rather, buried toxicity is closer to a “needle in a haystack”-like problem for raters. For example,~\citet{almerekhi2022investigating} found that identifying toxicity from comments in Reddit posts might be more difficult for human raters than comments on Twitter because the longer word limits allowed on Reddit make it challenging for raters to identify buried toxicity.\\ 


\textbf{Quotes or References to Toxic Content.} This refers to comments that quote or characterize the toxic speech of others for commentary or criticism, and yet, these comments sometimes end up getting unfairly flagged as toxic. This happens because such comments with quotes/references to toxic content represent a gray area for raters. Some raters may feel that such comments should be annotated as toxic regardless of whether the commentary associated with this quoted comment is toxic or not, because they do contain text which is toxic. The alternative perspective is that such comments need not be deemed toxic, as they may represent an individual pushing back against toxic speech by others. These two differing perspectives lead to rater disagreements.~\citet{van-aken-etal-2018-challenges} pointed out that many comments which quoted toxic references were annotated by raters as toxic or hateful, but in fact, they were non-hateful comments. Similarly,~\citet{risch2020toxic} enumerated examples of false positives generated by a toxicity detection system, i.e., otherwise non-toxic comments that cite toxic comments. As shown in Figure \ref{fig:venn_diagram}, this subtype of toxicity overlaps with \textit{buried toxicity}, as some of these quoted tweets may be quite long.

\subsection{Reasons based on Unclear Task Descriptions and/or Instructions}
\label{sec:Unclear task description}
In addition to reasons for disagreement that focus on raters (Section \ref{sec:heterogeneity}) or the comments themselves (Section \ref{sec:comment}), reasons based on unclear task descriptions could also lead to rater disagreement. In particular, we focus on the disagreement generated during the interactive process between job requesters and actual raters. For example, there could be issues related to the task description or instructions that are provided by job requesters to raters, or there could be issues related to the interpretation and understanding of these instructions (and overall context) by raters. Below, we discuss five subtypes of reasons for rater disagreement.\\

\textbf{Template Misunderstandings.} Disagreements are generated among raters as they might misunderstand the template (which refers to the description and instructions for the annotation task) and incorrectly categorize comments into toxicity. Without clear explanations and examples in the introduction of the template, crowd workers might misunderstand scientific visualizations which could be relevant for many annotation tasks~\citep{Yang_2023}. Further, an inability to communicate in real-time with the job requester to clear any doubts about the template is also a reason for misunderstanding and disagreement among raters. As a result, they have to re-annotate quite a few tasks due to unclear instructions~\citep{berg2015income}.\\

\textbf{Template Ambiguity.} This scenario describes comments for which the template doesn’t define what a correct choice should be in a given scenario. ~\citet{aroyo2015truth} indicated that the perceived problem of low rater agreement was due to undetailed and non-exhaustive guidelines for the annotation process. In fact, several job requesters do a very poor job of creating templates for their annotation tasks, e.g., templates which only provide the possible labels (or annotations) that raters are allowed to choose from, with no further instructions on how to interpret these labels, are commonly observed~\citep{poletto2021resources}.\\




\textbf{Text Ambiguity due to Missing Context.} A special case of template ambiguity (see Figure \ref{fig:venn_diagram}), this situation indicates comments that are difficult to understand without background context, which is often not provided to raters. For example, ~\citet{prabhakaran-etal-2020-online} pointed out that the annotation of a piece of text as hateful (or not) depended on the underlying context in which the text was written. Similarly,~\citet{yu2022hate} indicated that human judgments changed for most comments, when we show raters additional background context.\\

\textbf{Differences in Inferred Context.}
In situations where important context is missing, human raters are forced to infer the context for some of these cases, but they may do so in different ways. For example, if raters view a comment as a response to another toxic comment, they might not consider the comment toxic when placed in that context. ~\citet{wiebe2005annotating,fuoli2018stepwise}, and~\citet{fuoli2015optimising} indicated that the contextual nature of the annotations made the annotated data valuable for studying ambiguities that arose among subjective languages. As shown in Figure \ref{fig:venn_diagram}, this is a subtype of \textit{template ambiguity} as differences in inferred context arise due to ambiguous instructions provided as part of the annotation task template.\\


\textbf{Counterspeech.} Finally, this scenario indicates comments that counter speech/comments made by someone else (or counterspeech). Such counterspeech comments may not be viewed as toxic by some raters, while they might be viewed as toxic by others.~\citet{ruths2016counterspeech} highlighted the difficulty of determining the toxicity of counterspeech tweets/comments on Twitter. As shown in Figure \ref{fig:venn_diagram}, this sub-category overlaps with \textit{differences in inferred context}, \textit{template ambiguity}, and \textit{quotes/references to toxic content} (Section 3.2.2).

\subsection{Reasons related to randomness}
Finally, we discuss additional reasons for inter-rater disagreement that arise out of randomness in the crowdsourcing setup and annotation procedure. Since these reasons arise ``truly" at random, disagreement due to these reasons can be avoided to some extent.\\


\textbf{Variation at the Single Rater Level.} For borderline-toxic comments, repeatedly showing them to the same rater doesn’t necessarily yield consistent results, i.e.\ p(toxic | comment, rater) isn’t guaranteed to be binary even for a given comment and rater. This partly reflects limitations of using Likert scales to discretize our annotations: we don’t have a way to signal a slightly toxic comment, and such comments can slip under the radar. To the best of our knowledge, no existing work focuses on rater disagreements that arise out of this type of variation.\\

\textbf{Stochastic Rater Errors/Noise.} Rater errors or disagreements could also be caused by mistakes made by raters during annotation, e.g., misclicks, carelessness of raters, etc. In general, these stochastic errors/noise are easy to explicitly model within machine learning based toxicity detection systems. However, it is not easy to distinguish this source of disagreements from other more intractable sources of disagreement covered above. A long line of literature builds upon the Dawid-Skene model ~\citep{dawid1979maximum}, which provides an elegant computational framework that takes into account the expertise of different raters. It would be quite feasible and reasonable to extend such a model by incorporating (stochastic) rates with which individual raters may make mistakes in their annotations.\\

\textbf{Bad Crowdsourcing Practice.} Disagreements among raters also arises due to a lack of quality control measures employed by job requesters. ~\citet{kritikos2013survey,daniel2018quality} confirmed that many existing crowdsourcing platforms were not robust to effectively check and control the quality of crowdsourced tasks, which led to rater disagreements. As shown in Figure \ref{fig:venn_diagram}, this sub-category overlaps with \textit{template ambiguity}.

\section{Potential Solutions to Address Rater Disagreements}
\label{sec:solutions}
In this section, we summarize existing research that attempts to provide potential solutions for achieving high inter-rater agreement. Most of these existing approaches focuses on one or multiple reasons for disagreement that we have outlined as part of our taxonomy in Figure \ref{fig:comprehensive_taxonomy}. At a high level, most prior work has led to solutions of two kinds: (i) methods that eliminate or resolve rater disagreements to directly improve the annotation quality; and (ii) methods which treat disagreements as useful signals and utilize these disagreements to further improve the annotation accuracy. We discuss both kinds of solutions in this section.


\subsection{Resolving Rater Disagreement to Improve Annotation Quality}
\label{sec:solutions_eliminate}
Disagreements are frequently addressed as “error” or “noise”~\citep{kairam2016parting}. Thus, mainstream approaches aim their design to decrease and eliminate these disagreements~\cite{aroyo2019crowdsourcing}. Prior work in this area has been done separately (for the most part) by both the AI community and the crowdsourcing (and HCI) community. We discuss work done by each of these communities separately.\\

\noindent\textit{4.1.1 AI Approaches for Resolving Disagreements}

\noindent\textbf{Disagreements Related to Rater Heterogeneity.}~\citet{mozafari2020hate} proposed a bias alleviation mechanism to mitigate the effect of racial bias in the training set during the fine-tuning of the pre-trained BERT-based model for hate speech detection. In order to detect toxicity triggers (i.e., the turning points which make conversations toxic),~\citet{almerekhi2022provoke} created a dataset from Reddit and built a dual embedding biLSTM neural network with a set of sentiment shift, topical shift, and context-based features. Similarly,~\citet{10.1145/3342220.3344933} detected toxicity triggers by an LSTM neural network model using a combination of topical features, sentiment shift features, and GloVe embeddings.\\

\noindent \textbf{Disagreements Related to Issues with Comments.}~\citet{han-tsvetkov-2020-fortifying} proposed a framework to enhance toxicity classifiers against veiled toxicity through a few labeled probing examples.~\citet{kurita2019robust} developed the Contextual Denoising Autoencoder (CDAE) to learn robust representations based on character-level and contextual information to denoise obfuscating toxic tokens. Similarly, ~\citet{mishra2018neural} constructed a character-based word composition model that could encode obfuscated words. The integration of this composition model with an enhanced Recurrent Neural Network (RNN) model significantly advanced the state-of-the-art in the task of abuse detection. In order to solve the code-switching problem among comments,~\citet{roy2021leveraging} leveraged state-of-the-art Transformer language models to identify hate speech in a multilingual setting.~\citet{xie2022ensemble} presented another approach to deal with the code-switching problem and built toxicity models by applying the Jigsaw multilingual toxic comment classification dataset.~\citet{lees2022new} proposed a multilingual token-free Charformer model to detect potentially hateful and toxic messages. This approach extends the Perspective API into a more effective model across a diverse range of languages, usages, and styles, and performs especially well in code-switching, covert toxicity, emoji-based hate, and human-readable obfuscation.\\

\noindent \textbf{Disagreements Related to Unclear Task Descriptions and Instructions.} A lot of prior work attempts to discuss and eliminate rater disagreements that arise from unclear task descriptions or instructions by providing additional textual context.~\citet{menini2021abuse} investigated the role of textual context in abusive language detection based on Twitter datasets. They recommended that the textual context of messages should also be provided to raters since the context was sometimes necessary to understand the real intent of the user.~\citet{markov2022role} focused on detecting
the target of hate speech in Dutch social media and showed that the performance of the hate speech detection model could be significantly improved by integrating relevant contextual information.~\citet{10.1145/3561390} pointed out that raters changed their toxicity judgments in the presence of context. The structure of the context plays an important role during the annotation process.~\citet{xenos2021toxicity} created a dataset to further analyze the toxicity detection task from the perspective of conversational context. They found that if the conversational context is also considered, the practical quality of machine learning systems on the toxicity detection task can be further improved.\\

\noindent \textbf{Disagreements Related to Randomness based Reasons.}~\citet{raykar2010learning} proposed a probabilistic framework for supervised learning with multiple raters providing labels (but no absolute gold standard) to iteratively establish a particular gold standard, measure the rater performance given that gold standard, and then refine the gold standard based on the performance measures.~\citet{zhou2012learning} assumed that labels were generated by a probability distribution over workers, items, and labels. They proposed a novel minimax entropy principle to jointly estimate the distributions and the ground truth given the observed labels by workers.~\citet{wauthier2011bayesian} presented Bayesian Bias Mitigation for Crowdsourcing (BBMC), a Bayesian model to unify all three steps of crowdsourcing. This model captures raters' bias through a flexible latent feature model and conceives of the all three steps of crowdsourcing in terms of probabilistic inference.~\citet{chen2013pairwise} proposed a new model to predict a gold-standard ranking that hinged on combining pairwise comparisons via crowdsourcing. This approach takes into consideration the quality of the contributions of each rater.~\citet{ipeirotis2014repeated} formally analyzed the impact of repeated-labeling for data when
the labeling was found to be imperfect. This paper showed that under a wide range of conditions, repeated-labeling could improve both the quality of the labeled data directly, and the quality of the models
learned from the data.~\citet{zhang2014spectral}  proposed a two-stage efficient algorithm for multi-class crowd labeling problems that inferred the true labels from the noisy labels provided by non-expert crowdsourcing workers.~\citet{vempaty2014reliable} designed crowdsourcing systems through error-control codes and decoding algorithms for reliable
classification, despite the presence of unreliable crowd workers for the annotation task. The paper shows that the usage of good codes may improve the performance of the crowdsourcing task over typical majority-voting approaches.~\citet{yang2023semisupervised} proposed a semi-supervised classification algorithm based on noise
learning theory and a disagreement cotraining framework to reduce noise during the procedure of exchanging high-confidence samples among multiple models.\\

\noindent\textit{4.1.2 Crowdsourcing Approaches for Resolving Disagreements}

\noindent\textbf{Disagreements Related to Rater Heterogeneity.} As mentioned earlier, a long line of literature builds upon the Dawid-Skene model ~\citep{dawid1979maximum}, which provides an elegant computational framework that takes into account the expertise of different raters. For example,~\citet{5543189} proposed a model of the labeling process which included label uncertainty, as well as a multi-dimensional measure of the requester’s ability to estimate the reliability or expertise of the rater.\\

\noindent \textbf{Disagreements Related to Issues with Comments.} To the best of our knowledge, there is no prior work in the crowdsourcing literature on resolving differences arising from direct or indirect issues with comments.\\ 

\noindent \textbf{Disagreements Related to Unclear Task Descriptions and Instructions.} Turkopticon is an activist system that allows raters to publicize and evaluate their relationships with employers (or job requesters)~\citep{10.1145/2470654.2470742}. To a great extent, ratings on Turkopticon enable raters to put pressure on job requesters to set up better, and more informative annotation tasks for them, while also enabling raters to alert fellow raters about malicious or rude job requesters. In addition, Turker Nation is a general forum for Amazon Mechanical Turk (AMT) users, whose key function is to help reduce the information deficit and promote better collective action~\citep{martin2014being}.\\

\noindent \textbf{Disagreements Related to Randomness based Reasons.} In order to correct for the bias of raters (each of whom has a different level of expertise) who assign labels without actually looking at the comments,~\citet{raykar2011ranking} defined a score which can be used to rank the performance of raters.  ~\citet{rzeszotarski2011instrumenting} proposed a technique which captured behavioral traces from online crowd workers and used them to predict outcome measures (such quality, errors, and the likelihood of cheating) to reduce low-quality annotation.~\citet{ipeirotis2010quality} developed an algorithm to generate a scalar score representing the inherent quality of each worker. This score separates the intrinsic error rate from the bias of the worker, allowing for more reliable annotation quality estimation. Orthogonally,~\citet{snow2008cheap} proposed a bias correction method for non-expert raters to enhance annotation quality in the face of errors generated by low-quality workers.~\citet{alonso2014crowdsourcing} developed a framework to debug the design and template of subjective labeling tasks with low inter-rater agreement, and improve label quality before the crowdsourcing task runs at scale.

\subsection{Utilizing Rater Disagreements to Improve Annotation Quality}
Averaging the annotations received by multiple raters (i.e., taking the majority vote) has proven to be an effective strategy for incorporating rater disagreements into a single annotation, especially when a higher percentage of raters agree on a single annotation \cite{balayn2018characterising}. However, disagreement among human annotations is not necessarily noise, and hence it does not always have to be eliminated (e.g., by aggregating via majority vote). In fact, in a wide variety of domains, capturing disagreements among human annotations is crucial, as they represent plausible ranges of human judgments for some annotation tasks, where it is infeasible to define unique ground truth labels~\citep{pavlick-kwiatkowski-2019-inherent,47111,fleisig2023majority}. In such domains, aggregation (via majority vote and other such approaches) obscures crucial differences in opinion among raters, and hence it may even be counterproductive to eliminate disagreements (via aggregation) in these scenarios~\citep{fleisig2023majority}.

There is some prior work that discusses the importance of harnessing disagreements to enrich our understanding of the problem domain. For example,~\citet{aroyo2019crowdsourcing} indicated the importance of studying disagreements between raters and harnessing it to improve the quality of the crowdsourced data. A lot of other work focuses on using rater disagreement as a type of extra information.~\citet{gordon2022jury} introduced a supervised machine learning approach to replace the majority verdict with an opinion based on a ``jury''. This approach explicitly resolves disagreements enabling practitioners to make explicit value judgments.~\citet{fleisig2023majority} discussed the scenario of hate speech detection and pointed out that rater disagreements often resulted from differences among populations. They constructed a model that predicted individual annotator ratings on potentially offensive text. The model combines this information with the predicted target group of text to further predict the opinion of target group members on the offensiveness of the example. This is pretty helpful especially when they disagree with the raters in the majority group.~\citet{davani-etal-2022-dealing} introduced multi-annotator architectures to preserve and model the internal consistency in each rater’s labels as well as their systematic disagreements with other raters. They obtained better performance in modeling each rater as well as matching the majority vote prediction performance across different subjective classification tasks. 

Other work focuses on utilizing rater disagreements to make evaluations among raters or labels and control the quality of annotations.~\citet{abercrombie2023consistency} utilized rater agreements to measure label stability over time in order to provide quality control and provide insights about why raters disagree.~\citet{chang2017revolt} proposed Revolt, a collaborative approach that eliminated the burden of creating detailed label guidelines by harnessing crowd disagreements to identify ambiguous concepts and create rich structures for post-hoc label decisions.~\citet{aroyo2014three} introduced ``disagreement-based quality metrics'' for the annotation task of relation extraction. It illustrated how the disagreement in crowdsourcing could be harnessed to improve the quality of annotation data.~\citet{kairam2016parting} introduced the workflow design pattern of crowd parting and utilized disagreements to identify divergent, but valid, worker interpretations in crowdsourced entity annotations collected over two distinct corpora – Wikipedia articles and tweets on Twitter.

\section{Open Questions}
\label{sec:open_question}
In this section, we discuss some remaining gaps in the field of online toxicity research after giving our detailed taxonomy for the source of rater disagreements (Figure \ref{fig:comprehensive_taxonomy}), and call for corresponding solutions based on existing work. \\

\textbf{Developing Sophisticated Crowdsourcing Platforms. } Based on our discussion in Section 3.1.1, there are several issues with existing crowdsourcing platforms that make it difficult for job requesters to identify and recruit subject matter experts (or ideal raters). Note that we had mentioned that the list of premium qualifications for rater recruitment available on Amazon Mechanical Turk is not exhaustive. Even though it may be tempting to think of a future where this list of premium qualifications on Amazon Mechanical Turk (i.e., custom filters for sieving out non-ideal raters) grows large enough to accommodate most real-world scenarios, there are significant ethical considerations that counterbalance this discussion, as these qualifications could potentially be weaponized by malicious job requesters, and it could be used to discriminate against vulnerable raters belonging to marginalized and minority groups. 

At the same time, crowdsourcing platforms need to develop more robust protocols for conducting background and identity checks of raters (when they register as a crowd worker), so that issues such as intentional misrepresentation of identities can perhaps be mitigated. Similarly, in domains where rater beliefs evolve dynamically, it is important for crowdsourcing platforms to develop and adopt re-annotation protocols that enables the re-annotation of datasets periodically, so as to make these datasets more relevant to current-day societies.\\

\textbf{More Attention on Individual Differences.} All prior literature on rater disagreements has summarized the raters' behavior or subjectivity as belonging to one of the top categories of rater disagreements in their taxonomy definition~\citep{basile-etal-2021-need,jiang2022investigating,uma2021learning,sandri2023don}. As we discussed in Section~\ref{sec:heterogeneity}, we attempt to include a more detailed analysis of the rater's behaviors and discuss more subtypes of subjectivity from two aspects: recruitment factors and individual differences. The rater's behavior described in all previous literature belongs to recruitment-related factors (one of the categories defined in this paper) as prior work has focused its attention on how the rater's background, identity (demographic information), experiences, and task expertise influence the outcome of the toxicity annotation. However, how to evaluate the influence of the rater's individual preferences, beliefs, and opinions towards toxicity annotation deserves further discussion. What's more, how to design the annotation procedure and guide raters to reduce the levels of annotation disagreement (taking into account that rater beliefs always change over time) also remains an open problem. \\

\textbf{Moving on from "True" Randomness to Next Steps.} In the past, disagreement has been thought of as the product of the rater's errors (or "true" randomness defined in this paper), i.e., all the combined mistakes or slips made by the rater.~\citet{nedoluzhko-etal-2016-coreference} also indicated that a high proportion of rater disagreements are generated due to "true" randomness. Aiming to solve this, the crowdsourcing community has put a lot of effort in understanding how to evaluate the reliability of the rater and improve annotation quality. Main directions of toxicity annotation or natural language processing annotations focus on filtering low-quality raters and annotations or aggregating labels among different raters through majority voting. However, as more works investigate different alternative reasons that lead to rater disagreements in the toxicity annotation domain ( in Figure \ref{fig:comprehensive_taxonomy}) , the corresponding approaches for handling other types of disagreements need to be extended by the crowdsourcing community as well.\\

\textbf{Utilizing Rater Disagreements.} Despite our best efforts, we found limited literature on utilizing and harnessing disagreements to improve the annotation quality. Existing approaches mainly focus on the procedure of label aggregation (through majority voting) or discuss ways for evaluating the quality of labels or the qualification of raters. Among these approaches,~\citet{fleisig2023majority} is the only one which utilizes rater disagreement while taking consideration of the raters' heterogeneity (i.e.\ demographic information) at the same time. How to utilize disagreement while incorporating other reasons (discussed in Section~\ref{sec:reasons}) remains an exciting and open area of research.\\

\textbf{Toxicity on multi-lingual text.} Almost all literature in this survey discusses rater disagreements of toxicity annotation based on English language text since it is the majority language that is used on the Internet. Even though we argue that the multi-lingual nature of several texts is one of the many reasons leading to rater disagreements, an extensive study of rater disagreement in multi-lingual and code-mixed settings is yet to be done. As a large proportion of people on the planet (especially people from the Global South) engage in code-mixed expression, code mixing is increasing becoming common in online communication, and thus, understanding the toxicity and providing corresponding solutions under this scenario deserves further discussion.

\section{Limitations}
\label{sec:limitation}
While the taxonomy of reasons which cause rater disagreements is proposed to cover the annotation of subjective tasks, we apply it to a smaller scope of online toxicity annotation. Therefore, the literature pertaining to disagreement reasons that appears in this survey is limited to the topics related to annotation, crowdsourcing, rater disagreements, and toxicity (including cyberbullying, hate speech, offensive content, and abusive language). In addition, while we cover several mitigation strategies in Section 4, several mitigation strategies that we cover in our article are not explicitly in the scope of toxicity detection. That being said, the mitigation strategies that we cover are general-purpose techniques which should readily generalize to the toxicity classification domain. Also, the taxonomy proposed in this survey is comprehensive, yet there might be some overlaps between the different reasons that we have outlined.


\section{Conclusion}
\label{sec:conclusion}
In this paper, we discuss the reasons behind rater disagreement with a special focus on the annotation of online toxicity. To the best of our knowledge, this is the first comprehensive work that analyzes different types of reasons. We develop a taxonomy based on these reasons and summarize corresponding solutions from existing work in artificial intelligence, and crowdsourcing communities. With this comprehensive taxonomy, we further identify corresponding restrictions, as well as potential areas of focus for the development of subsequent research.

\bibliographystyle{ACM-Reference-Format}
\bibliography{sample-base}










\end{document}